\documentclass[twoside,11pt]{article}

\usepackage[accepted]{melba}
\usepackage{xcolor}
\usepackage{mwe} 
\usepackage{comment}

\usepackage{amsmath,amsfonts}

\renewcommand{\textcolor}[2]{#2}

\melbaid{2024:030}  
\doi{https://doi.org/10.59275/j.melba.2024-267f}
\melbaauthors{Rabbi, Kiechle, Beaulieu, Ray and Cobzas}  
\volume{2}
\firstpageno{2268}  
\melbayear{2024}  
\datesubmitted{04/2024}  
\datepublished{11/2024}  

\melbaspecialissue{}
\melbaspecialissueeditors{}

\ShortHeadings{Autoencoder for Hippocampal Shape Variations}{Rabbi, Kiechle, Beaulieu, Ray and Cobzas}

\title{Disentangling Hippocampal Shape Variations: A Study of Neurological Disorders Using Mesh Variational Autoencoder with Contrastive Learning}

\author{\firstname Jakaria \surname Rabbi\orcid{0000-0001-9572-9010} \email jakaria@ualberta.ca \\  
	\addr Department of Computing Science, University of Alberta, Edmonton, Canada.
	\AND
	\name Johannes Kiechle 
	    \orcid{0009-0004-9610-146X}
	 \email johannes.kiechle@tum.de \\
	\addr Institute for Computational Imaging and AI in Medicine, Technical University of Munich, Munich, Germany.
 	\AND
	\name Christian Beaulieu\orcid{0000-0002-2342-3298} \email christian.beaulieu@ualberta.ca \\
	\addr Department of Radiology and Diagnostic Imaging \& Biomedical Engineering, University of Alberta, Edmonton, Canada.
 	\AND
	\name Nilanjan Ray\thanks{Equal contribution.
} \orcid{0000-0002-7588-5400} \email nray1@ualberta.ca \\
	\addr Department of Computing Science, University of Alberta, Edmonton, Canada.
  	\AND
	\name Dana Cobzas\footnotemark[1] \orcid{0000-0003-1318-4868} \email cobzasd@macewan.ca \\
	\addr Department of Computer Science, MacEwan University, Edmonton, Canada.
}

\begin{document}

\maketitle

\begin{abstract}
This paper presents a comprehensive study focused on disentangling hippocampal shape variations from diffusion tensor imaging (DTI) datasets within the context of neurological disorders. Leveraging a Mesh Variational Autoencoder (VAE) enhanced with Supervised Contrastive Learning, our approach aims to improve interpretability by disentangling two distinct latent variables corresponding to age and the presence of diseases. In our ablation study, we investigate a range of VAE architectures and contrastive loss functions, showcasing the enhanced disentanglement capabilities of our approach. This evaluation uses synthetic 3D torus mesh data and real 3D hippocampal mesh datasets derived from the DTI hippocampal dataset. Our supervised disentanglement model outperforms several state-of-the-art (SOTA) methods like attribute and guided VAEs in terms of disentanglement scores. Our model distinguishes between age groups and disease status in patients with Multiple Sclerosis (MS) using the hippocampus data. Our Mesh VAE with Supervised Contrastive Learning shows the volume changes of the hippocampus of MS populations at different ages, and the result is consistent with the current neuroimaging literature. This research provides valuable insights into the relationship between neurological disorder and hippocampal shape changes in different age groups of MS populations using a Mesh VAE with Supervised Contrastive loss.
	\newline Our code is available at~\url{https://github.com/Jakaria08/Explaining_Shape_Variability}.
\end{abstract}


\section{Introduction}
\subsection{Motivation and Objective}
Advances in shape analysis and disentanglement techniques have contributed significantly to medical imaging, particularly in the 2D and 3D analysis of anatomical structures \citep{intro_shape_analysis}. Latent space refers to a lower-dimensional representation of the complex high-dimensional space inherent in the data. Disentanglement involves the extraction and isolation of independent factors within this latent space, enabling a more interpretable and meaningful representation of anatomical variations from 2D and 3D datasets in the realm of medical imaging \citep{intro_shape_analysis_dis}. \par
Integrating latent space disentanglement techniques in medical image analysis helps reveal hidden factors that contribute to the observed variations in shapes and structures. Within the paradigm of 3D shape analysis, disentangling latent spaces holds profound potential for unraveling the complexities of diseases and age-related variations in brain structures \citep{intro_shape_analysis_age}. By isolating and understanding these latent factors, researchers can pave the way for more accurate diagnostic tools, predictive models, and a deeper comprehension of the underlying conditions driving anatomical changes. \par
Investigating the shape changes of the hippocampus over different age groups, especially in the context of neurological disorders, is complex when longitudinal data is not available, such as multiple Magnetic Resonance Imaging (MRI) scans taken at various ages for the same individual \citep{intro_shape_analysis_age}. Nevertheless, as shown through the current study, valuable information about hippocampal morphology and atrophy can still be extracted from existing limited datasets. Our proposed method intends to discern and describe the hippocampal shape variations in individuals across various age groups, differentiating between those with and without neurological conditions such as Multiple Sclerosis (MS) \citep{MS_hippocampal_change}. \par
We use 3D mesh representation for our experiments as opposed to images or point clouds \citep{intro_shape_analysis_age}. The advantages of 3D mesh representation include its ability to capture complex surface details, providing a high-fidelity representation for studying anatomical shape variability. This representation allows for more intuitive and interpretable shape analysis, directly examining surface geometry for a clearer understanding of morphological changes \citep{mesh_surface_detail}. The 3D mesh also facilitates the application of advanced shape analysis techniques, aligning well with methodologies like statistical shape models and deep learning. \par
We employ Mesh VAE to obtain interpretable latent dimensions and generate valid shapes similar to the training dataset. We use a modified contrastive loss \citep{SNN_contrastive_loss} as a latent space disentanglement strategy to isolate two data generative factors: age and disease (MS). Age is represented by continuous values, while disease is labeled by discrete values. This combination of regression and classification tasks is frequently encountered in medical imaging. Our model formulation results in more interpretable latent codes and enables control over the generative process based on the specified factors (both classification and regression). As part of our validation process for the proposed method, we develop a 3D synthetic torus dataset with four factors of variability. During training and testing, we disentangled two of these factors using labels. Our results demonstrate supervised disentanglement using both classification and regression data, both combined and separately.

\subsection{Related Works}
The disentanglement of the latent space has been the focus of numerous research works. In this section, we provide an overview of both supervised and unsupervised methods for disentangling latent variables. In section \ref{DLR VAE}, we discuss the vanilla VAE and related works that enhance the disentanglement performance of the vanilla VAE. Most of these methods are unsupervised and disentangle the latent space without any prior knowledge of which variable disentangles which data-generating factor. We also examine supervised VAEs that disentangle specific latent variables using the data labels, and these methods exhibit strong disentangling performance for specific factors when compared to unsupervised approaches. Furthermore, we present some contrastive learning-based methods that enhance disentanglement, although they are also unsupervised methods. Additionally, we explore some graph autoencoder methods. Finally, in section \ref{3d_mesh_dis}, we explore disentanglement techniques that employ 3D mesh data with self-supervision and conditional VAEs and are related to our proposed method.
\subsubsection{Disentangled Latent Representation VAEs}\label{DLR VAE}

Variational Autoencoders (VAEs) represent a powerful class of generative models in machine learning that aim to capture the underlying structure of complex data \citep{vae}. VAEs consist of an encoder network, which maps input data to a probability distribution in a latent space, and a decoder network that reconstructs the input from sampled points in that space. The concept of disentanglement in VAEs addresses the challenge of extracting interpretable and independent features from the latent representation. \par
\citet{betavae} introduced $\beta$-VAE, a framework for obtaining interpretable latent representations from raw image data through unsupervised learning. $\beta$-VAE modifies the traditional VAE by incorporating an adjustable hyperparameter, $\beta$, which influences the trade-off between latent channel capacity, independence constraints, and reconstruction accuracy. Factor VAE, introduced by \citet{factorvae}, addresses the issue of overly compact representations of $\beta$-VAE by incorporating a total correlation term in the objective function, promoting more independent and disentangled latent variables. DIP-VAE (Disentangled Inferred Prior VAE), proposed by \citet{DIPvae}, aims at mitigating the learning of trivial latent dimensions by introducing a penalty term that encourages the inferred posterior to have fixed marginals. Another method, $\beta$-TC-VAE ($\beta$-Total-Correlation-VAE), introduced by \citet{betatcvae}, dynamically adapted the $\beta$ hyperparameter for each latent dimension based on the total correlation, striking a balance between disentanglement and reconstruction accuracy. \par

Several methods provide insights and theoretical assessments on disentangled representations in VAEs, specifically in $\beta$-VAE. \citet{annealed-vae} proposed a training process modification for $\beta$-VAE that progressively increases latent code information capacity, facilitating robust learning of disentangled representations without sacrificing reconstruction accuracy. \citet{dava} proposed another training approach for variational auto-encoders named DAVA (Disentangling Adversarial Variational Autoencoder). DAVA effectively addresses the challenge of hyperparameter selection, reducing dependence on dataset-specific regularization strength. Additionally, \citet{jointvae} proposed an unsupervised framework for learning interpretable representations, combining continuous and discrete features using variational autoencoders.  \par

Some related works utilize supervised disentanglement methods, enhancing disentanglement by utilizing specific latent variables and labels. \citet{guidedvae} discussed methods employing both unsupervised and supervised learning in the context of generative models. They introduced an algorithm, Guided VAE, aimed at achieving controllable generative modeling through latent representation disentanglement learning. \citet{attributevae} proposed a supervised approach called Attri-VAE, which employs a VAE to generate interpretable representations of medical images. This method includes an attribute regularization term, associating clinical and medical imaging attributes with different dimensions in the latent space, facilitating a more disentangled interpretation of attributes. \par
An alternative approach for disentangling the latent space is to use contrastive learning in VAEs that proves beneficial for achieving improved disentanglement and generative performance. In a study by \citet{contrastive1}, a methodology is introduced to generate facial images of virtual individuals with controlled and disentangled latent representations for identity, expression, pose, and illumination. The contrastive learning strategy is also applied to train autoencoder priors \citep{contrastive2} and masked autoencoders \citep{contrastive3}.
\par
Another type of autoencoder called Graph autoencoder has become a crucial tool for studying graph-structured data. It enables the learning of meaningful representations for tasks such as node classification, link prediction, and clustering. A pioneering contribution to this domain was made by Kipf et al., who introduced the Graph Convolutional Network (GCN) in an autoencoder framework. Their approach, known as the Variational Graph Autoencoder (VGAE) \citep{vgae}, combines the power of GCNs to aggregate and propagate node features with a variational autoencoder’s capacity for learning interpretable latent representations. Further advancements include the work of  Pan et al.'s Adversarially Regularized Graph Autoencoder (ARGA), and Adversarially Regularized Variational Graph autoencoder (ARVGA) \citep{vgae2}, which incorporates adversarial training to enhance the robustness of learned representations, and Wang et al.'s Marginalized Graph Autoencoder (MGAE) \citep{wang2017mgae}, which introduced a denoising-based approach specifically tailored for graph clustering tasks.

\subsubsection{Disentanglement in 3D Mesh Data using Mesh VAEs}\label{3d_mesh_dis}
While most disentangling methods are designed for images, researchers also use self-supervised and conditional VAEs to disentangle specific attributes in 3D mesh datasets. One of the research \citep{3Dvaeswap} introduced a self-supervised method for training a 3D shape VAE aimed at achieving a disentangled latent representation of identity features in 3D generative models for faces and bodies. The approach involves mini-batch feature swapping between various shapes to optimize mini-batch generation and formulate a loss function based on known differences and similarities in latent representations. \citet{3dvaesecondface} proposed a VAE framework to disentangle identity and expression from 3D input faces that have a wide variety of expressions. \par
The two approaches mentioned for 3D VAE are not applicable to our specific medical domain problem, as we find neither feature swapping nor unsupervised learning appropriate. In our case, we possess labeled data and aim to disentangle multiple latent variables (for classification and regression) with supervision because supervised training increases the disentangling and reconstruction performance according to the previous research we discussed. Our work partially uses the process proposed by \citet{intro_shape_analysis_age}, who explore a supervised variational Mesh autoencoder to understand and explain the variability in anatomical shapes. However, they only use the excitation-inhibition mechanism for the regression problem and used two additional neural networks for their method, which is different from our proposed method. \par
The domain of medical imaging often necessitates the disentanglement of various factors, including but not limited to age, diseases, and gender, through both classification and regression techniques. Therefore, we focus on simultaneous classification and regression techniques in the medical imaging domain with better loss functions. Our decision to focus on the two latent factors (classification and regression tasks) in the hippocampal study was primarily guided by their strong biological relevance and interpretability, particularly in the context of neurodegenerative diseases. These factors align with well-understood morphological variations in hippocampal structures, making our model's output more relevant for clinical research. The availability of supervised labels for these specific dimensions further supports our choice, allowing us to validate the model’s disentanglement and ensure that the learned representations are meaningful and clinically significant. To the best of our knowledge, no prior work has utilized contrastive loss for both the classification and regression tasks (simultaneously) with a guided mesh VAEs to disentangle multiple latent variables. Our proposed method demonstrates superior disentanglement compared to guided VAE \citep{guidedvae} and attribute VAE \citep{attributevae} while achieving comparable generative quality and speed. 

\subsection{Contributions}
The main contributions of this paper are summarized as follows:
\begin{itemize}
    \item We introduce a novel contrastive loss for categorical and continuous labels to improve the disentanglement performance of specific latent variables through supervised learning using 3D mesh data.
    \item Our unified loss function incorporates both excitation and inhibition mechanisms for classification and regression tasks.
    \item We apply our novel formulation to analyze anatomical shape variations across various factors, including age and disease (MS), through the generation of 3D shapes.
\end{itemize}


\section{Materials and Methods}
Our proposed VAE designed for deep mesh convolution operates with an input comprising 3D mesh vertices denoted as $X = [x_{0}, x_{1}, . . . , x_{N-1}]^T \in \mathbb{R}^{N\times F}$, where $F$ represents the feature dimension, and $N$ is the total number of vertices per mesh. In the context of 3D mesh data, $F$ is specified as 3, and $X$ contains the coordinates of each vertex.

\begin{figure}[h!]
    \centering
    \includegraphics[width=\linewidth]{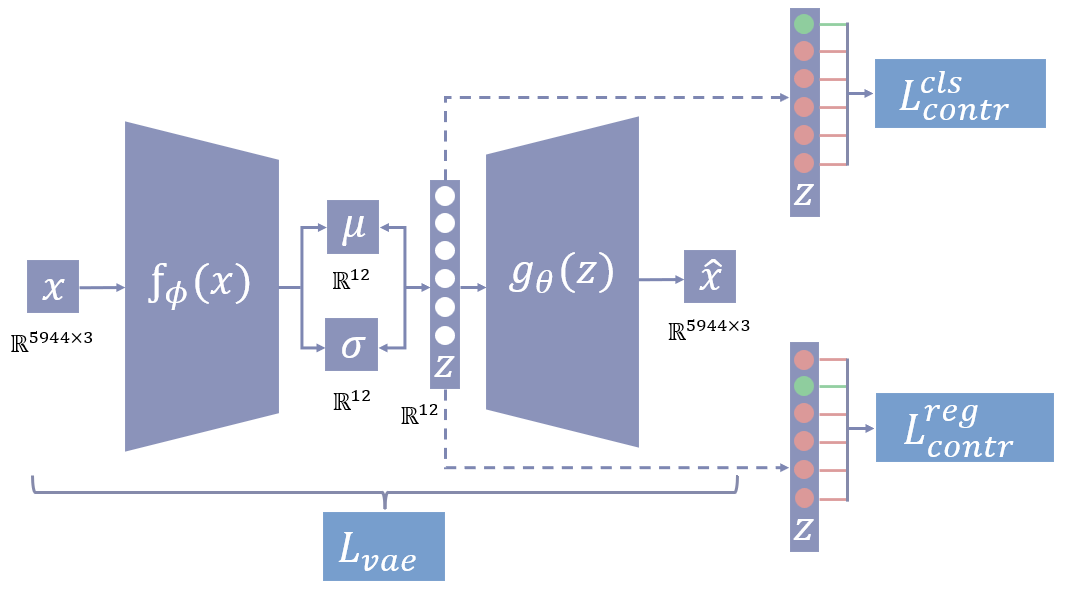}
\caption{Overall architecture of our method. We have mesh VAE with an encoder $f_\phi(x)$ and decoder $f_\theta(z)$ where $x$ is the input 3D mesh and $z$ represents the latent space. $L_{vae}$ represents the VAE loss that combines reconstruction and KL divergence loss. Another two losses are classification loss $L_{contr}^{cls}$ and regression loss $L_{contr}^{reg}$, where a specific latent variable is disentangled for a specific feature (continuous or discrete). We use the first variable for contrastive classification loss ($z_{1}$ corresponds to binary labels, and the rest variables are uncorrelated to the labels). The second variable $z_{2}$ corresponds to regression loss, and the rest variables are uncorrelated to the continuous labels. 
}
\label{fig:overall_archi}
\end{figure}

The network's encoder is based on the SpiralNet++ structure \citep{spiralnet++}, where all vertices of an input mesh are interconnected via a spiral trajectory that initiates from a randomly chosen vertex. The execution of spiral convolution operations involves two steps: initially, mesh vertices along the trajectory within a fixed distance are concatenated, a process known as neighborhood aggregation. Following this, the concatenated vertices undergo processing through a multilayer perceptron (MLP) with weight sharing \citep{intro_shape_analysis_age}. The decoder module performs a reverse transformation compared to the encoder using the latent space $z$. Our overall network architecture is shown in Figure \ref{fig:overall_archi}.
\par
Our method uses SpiralNet++ to exploit the local geometric structure of mesh data, preserving spatial relationships between vertices, which is crucial for accurately modeling the hippocampus and torus data. In contrast, we do not use PointNet-type models that focus on global features and may lose critical local geometric information. SpiralNet++ ensures consistent capture of local features through a fixed template, aligning with our goal of disentangling specific factors within hippocampus data \citep{spiralnet++}. The importance of maintaining local mesh structure for accurate shape analysis, as highlighted by Litany et al. \citep{litany2018deformable}, further justifies the use of SpiralNet++ over PointNet-type models.
\par
In the following section, we present the $\beta$-VAE used for our formulation. Then we discuss supervised guided VAE to explore the excitation-inhibition mechanism. However, we implement the mechanism differently. Finally, in section \ref{SCVAE}, we show the formulation of our method.

\subsection{$\beta$-VAE \citep{betavae}}
Our approach uses the $\beta$-VAE as the backbone of our network architecture. The VAE uses the Evidence Lower Bound (ELBO) as its objective function, expressed as:
\begin{equation}
    \max_{\theta,\phi}\{L_{ELBO}(\theta, \phi) = \mathbb{E}_{q_{\phi}(z | x)}[\log p_\theta(x | z)] - \beta \cdot KL(q_\phi(z | x) \| p(z))\}
\end{equation}
This equation reflects the balance between maximizing the reconstruction accuracy of the model, quantified by the expected log-likelihood of the observed data ($x$) given latent variables ($z$), which is $\log p_\theta(x | z)$, and minimizing the divergence between the posterior distribution of the latent variables under the encoder model $q_\phi(z | x)$ and the prior distribution $p(z)$, scaled by the hyperparameter $\beta$. Here, $\phi$ and $\theta$ are the parameters of the encoder and decoder network. This objective function can be decomposed into two primary components: the reconstruction loss and the Kullback-Leibler divergence, forming the VAE loss demonstrated by the following equations:

\begin{equation}
L_{vae}=L_{reconstruction}+\beta\cdot L_{KL}
\end{equation}
where, $L_{vae}$ is the overall $\beta$-VAE loss consisting of reconstruction and KL divergence loss (multiplied with hyperparameter $\beta$).
\begin{equation}
L_{reconstruction} = \| \hat{X} - X \|^{2}
\end{equation}
where, $X$ is the input 3D mesh shape and $\hat{X}$ is the reconstructed shape.
\begin{equation}
   L_{KL}= KL(q_\phi(z | x) \| p(z))
\end{equation}
where, $L_{KL}$ measures the KL divergence between the posterior ($q_\phi(z | x)$) and prior ($p(z)$) distributon.
\subsection{Supervised Guided VAE \citep{guidedvae}}
The objective function of the supervised guided VAE model is defined as follows:
\begin{equation}
    \max_{\theta,\phi}\{L_{ELBO}(\theta, \phi) + L_{Excitation}(\phi, t) - L_{Inhibition}(\phi, t)\}
\end{equation}
where $L_{ELBO}$ is the evidence lower bound, $L_{Excitation}$ is the excitation loss, and $L_{Inhibition}$ is the inhibition loss calculated using separate feed-forward neural networks. The excitation loss is used to establish a correlation between a specific latent variable, for example, $z_{1}$, and a data generative factor like age or scale.  The excitation loss is defined as:
\begin{equation}
    L_{Excitation}(\phi, t) = \max_{\omega} \{E_{q_{\phi}(z_t|x)}[\log p_{\omega}(y|z_t)]\}
\end{equation}
where $z_{t}$ is the latent variable for supervised disentanglement, $x$ is the input data, $y$ is the label, and $\omega$ parameterizes the excitation network. The inhibition loss is defined as:
\begin{equation}
    L_{Inhibition}(\phi, t) = \max_{\gamma} \{E_{q_{\phi}(z_{-t}|x)}[\log p_{\gamma}(y|z_{-t})]\}
\end{equation}
In this context, $z_{-t}$ represents a composite of latent variables excluding $z_t$, and $\gamma$ denotes the parameters of the inhibition network. The methodology involves the training of distinct latent variables ($z_{t}$) to establish correlations with specific features within a dataset ($y$). The inhibition term is designed to avoid unintended associations between other latent variables ($z_{-t}$) and the labeled output. Our approach follows the excitation-inhibition mechanism like the guided VAE but with different losses without the need for separate neural networks. 
\subsection{Supervised Contrastive VAE (Ours)} \label{SCVAE}
We introduce a contrastive loss based on \citet{SNN_contrastive_loss} applied to the excitation-inhibition mechanism inspired by \citet{guidedvae} and \citet{intro_shape_analysis_age} including a threshold hyperparameter in the regression loss for disentangling latent space of a VAE. The loss function $L_{contr}$ is composed of three parts: $L_{vae}$, $L_{contr}^{cls}$, and $L_{contr}^{reg}$ shown in \ref{eq:8}. The first part is the loss function of the VAE, while the other two are the contrastive loss functions. 
\par Contrastive loss learns the representations of the data in the latent space, and our first contrastive loss function $L_{contr}^{cls}$ enforces the similarity between the samples of the same class and the dissimilarity between the samples of different classes. \textcolor{blue}{In our problem setting, classification inherently deals with discrete labels, where the model's objective is to distinguish between distinct classes. The loss function $L_{contr}^{cls}$ encourages the latent space to separate data points based on these discrete labels. This loss only applies to the latent variable responsible for the classification task.} We use it to disentangle the first latent variable ($z_{1}$) that correlates with the bump (present or absent) in the torus dataset and disease (healthy or MS) in the hippocampus dataset.  Here, the binary labels are represented by $y$. The loss is demonstrated in Figure \ref{fig:overall_archi}.
\begin{equation}
    L_{contr}=L_{vae} + L_{contr}^{cls} + L_{contr}^{reg}
    \label{eq:8}
\end{equation}

\begin{equation}
L_{contr}^{cls} = -\frac{1}{b} \sum_{i\in1..b} \log \left( \frac{\sum_{\substack{j\in1..b \\ j\neq i \\ y_i=y_j}} e^{-\frac{||z_{1}^i-z_{1}^j||^2}{T}}}{\lambda_{1}\sum_{\substack{k\in1..b \\ k\neq i}} e^{-\frac{||z_{1}^i-z_{1}^k||^2}{T}} + \lambda_{2}\sum_{\substack{k\in1..b \\ k\neq i \\ y_i=y_k}} e^{-\frac{\sum_{d\in2..d_{z}}||z_{d}^i-z_{d}^k||^2}{(d-1)T}} } \right)
\label{eq:9}
\end{equation}

In equation \ref{eq:9}, the loss function is estimated across the data batch $b$ by sampling a neighboring point $z_{1}^j$ for each point $z_{1}^i$ in the latent space. The likelihood of sampling $z_{1}^j$ depends on the distance between points $z_{1}^i$ and $z_{1}^j$. The loss is represented by the negative logarithm of the probability of sampling a neighboring point $z_{1}^j$ from the same class ($y$) as $z_{1}^i$. The temperature parameter $T$ regulates the significance assigned to the distances between pairs of points. We implement the inhibition mechanism by introducing a term $\sum_{\substack{k\in1..b \\ k\neq i \\ y_i=y_k}} e^{-\frac{\sum_{d\in2..d_{z}}||z_{d}^i-z_{d}^k||^2}{(d-1)T}}$, weighted by $\lambda_{2}$ in the denominator of the loss and the formulation ensures that other latent variables from $z_{2}$ to $z_{d_z}$ ($d_z=$ number of latent variables) remain uncorrelated with the classification labels. We use all the values from the other dimensions from the latent space within the exponential term and use $d-1$ in the denominator to take the average. The loss formulation acts like excitation unit when $\lambda_{2}=0$ and $\lambda_1=1$.  \par

\textcolor{blue}{On the other hand, Regression deals with continuous targets in our setting. We employ a threshold parameter (Th) in equation \ref{eq:10} to simulate a classification problem using continuous targets. This is a hyperparameter that controls the granularity of the regression task. It divides the values into separate bins and is completely different from the classification loss function. This loss only applies to the latent variable responsible for the regression task.} We formulate the loss function to address our regression problem of disentangling $z_{2}$ (depicted in figure \ref{fig:overall_archi}) based on continuous labels such as ages or scales. The loss function categorizes data objects into the same class if their labels fall within a specified range ($|y_i-y_j|\leq Th$) determined by the threshold.
\begin{equation}
L_{contr}^{reg} = -\frac{1}{b} \sum_{i\in1..b} \log \left( \frac{\sum_{\substack{j\in1..b \\ j\neq i \\ |y_i-y_j|\leq Th}} e^{-\frac{||z_{2}^i-z_{2}^j||^2}{T}}}{\lambda_{1}\sum_{\substack{k\in1..b \\ k\neq i}} e^{-\frac{||z_{2}^i-z_{2}^k||^2}{T}} + \lambda_{2}\sum_{\substack{j\in1..b \\ j\neq i \\ |y_i-y_j|\leq Th}} e^{-\frac{\sum_{d\in{1,3..d_{z}}}||z_{d}^i-z_{d}^j||^2}{(d-1)T}}} \right)
\label{eq:10}
\end{equation}
\par
Overall, we employ the modified soft nearest neighbor losses (SNNL) \citep{SNN_contrastive_loss} as part of the inhibition-excitation mechanism, aiming to disentangle specific latent variables associated with distinct data generative factors. Our approach is scalable, allowing extension to more than two latent variables to disentangle additional data-generative factors of interest. While our model can disentangle other latent variables except the targetted ones, it does not enforce supervision for those variables. The  SNNL focuses on enhancing latent representations' quality by promoting similarity among embeddings and assigning probabilities to all samples.
\par 
There exists an alternate contrastive loss, InfoNCE \citep{infonce}, which is formulated as a binary classification task, distinguishing positive from negative pairs. It compels the model to learn representations where positive pairs are more similar to each other than to negative pairs, effectively maximizing mutual information between positive pairs. However, we opt for a modified SNNL due to its less explicit differentiation between positive and negative pairs, emphasizing the creation of a smoother, probabilistic representation of similarity. 
\par
The denominator of our modified SNNL (equation \ref{eq:9} and \ref{eq:10}) involves a sum over the exponential terms of all latent representations ($z_{1}$ and $z_{2}$) of the samples in the dataset, encompassing both positive and negative samples, and it encourages the model to assign higher probabilities to positive pairs without enforcing a strict binary distinction, as InfoNCE does. Therefore, our model and loss function can address both classification (equation \ref{eq:9}) and regression problems (equation \ref{eq:10}) using SNNL. The inclusion of an extra term in the denominator, weighted by $\lambda_{2}$, enhances the probability of attaining a more disentangled representation. This is achieved by considering all latent representations in the variables not intended for disentanglement for a specific data generative factor.

\section{Experiments and Analysis}
\subsection{Datasets}
In this section, we provide an overview of the datasets utilized in this study, namely the hippocampus and synthetic data. All models compared in the results section are assessed using the data from both datasets.
\subsubsection{Synthetic Torus Dataset}
\begin{figure}[h!]
    \centering
    \includegraphics[width=\linewidth]{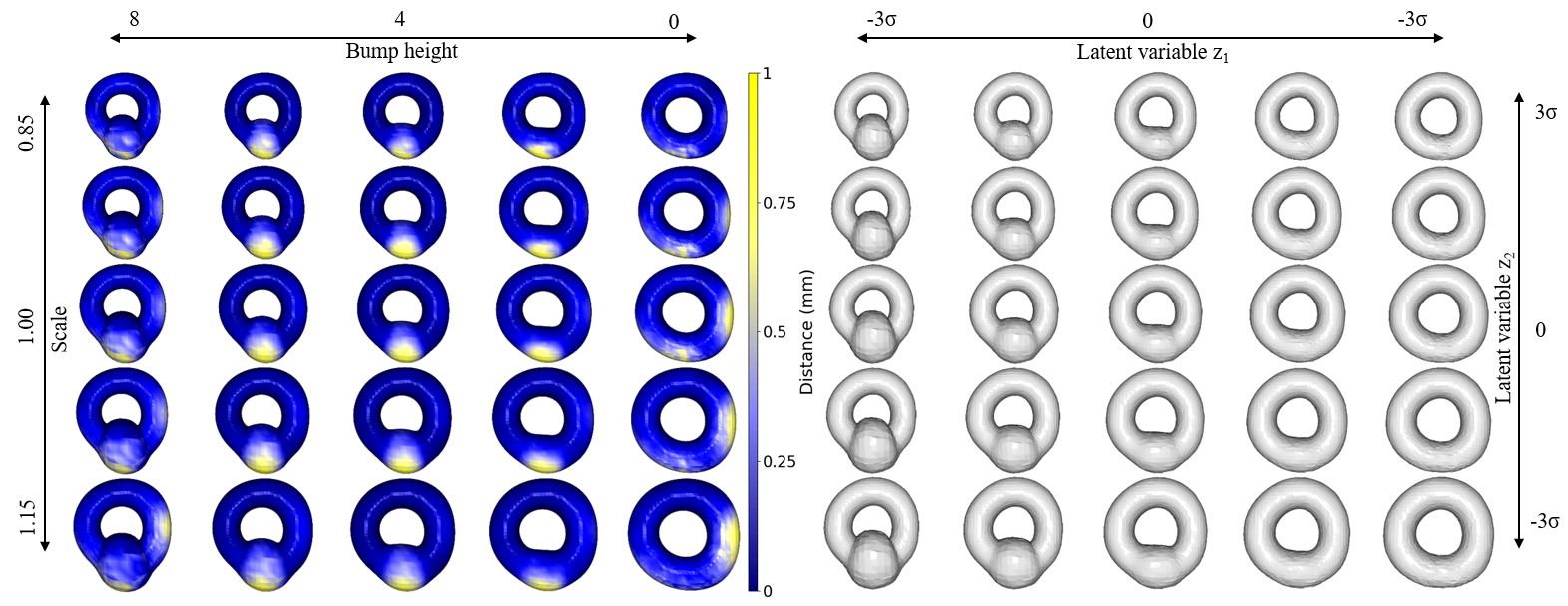}
    \caption{On the left side, we show the combination of reconstructions and original torus meshes from the synthetic
dataset using our proposed model. The dark blue indicates a very small deviation between the reconstruction and the original mesh. We show two variabilities in the matrix of images: bump height and scale. On the right side, we show the decoder’s output by varying the disentangled latent variable $z_{1}$ and $z_{2}$ in the x and y axis while holding the other latent variables constant at a mean value which is zero.
    }
\label{fig:torus_dataset_vis}
\end{figure}

We have a hippocampus dataset that only includes a single scan per subject, it lacks the necessary ground truth to establish the relationship between shape and age. Furthermore, the data only offers scans for healthy and MS populations separately. Longitudinal data, on the other hand, can provide insight into the hippocampal shape of individual subjects over time, taking into account their MS status. Consequently, for evaluation purposes, synthetic data representing a torus with a bump (varying in size and presence/absence) is utilized following the method introduced in an article by \citet{intro_shape_analysis_age}. We introduce four types of variability (scale of the torus, different noises, presence, and height of the bump) but only two variabilities (bump presence and torus scale) are disentangled in the latent space. We generate 5000 torus data by varying generative factors for our experiments. In figure \ref{fig:torus_dataset_vis}, the color difference illustrates the dissimilarity between original and generated torus shapes, highlighting the variations in torus shapes by adjusting the values of the two latent variables controlling torus bump size and total scale.

\subsubsection{Hippocampus Dataset}
We utilize a neuroimaging dataset that incorporates diffusion tensor imaging (DTI) scans. The high-resolution data displays a voxel size of 1 mm isotropic, is acquired at 3 Tesla, and consists of volumes measuring 220 × 216 × 20 mm³ \citep{hippocampus_data_acquisition}. This dataset encompasses scans from 204 healthy subjects spanning an age range between 32 to 71 years, with 112 females. Additionally, we have scans from subjects with MS (43 subjects aged between 32 to 71, with 35 females and the rest being males) \citep{MS_hippocampal_change}.
\begin{figure}[h!]
    \centering
    \includegraphics[width=5 in, height=1.7 in]{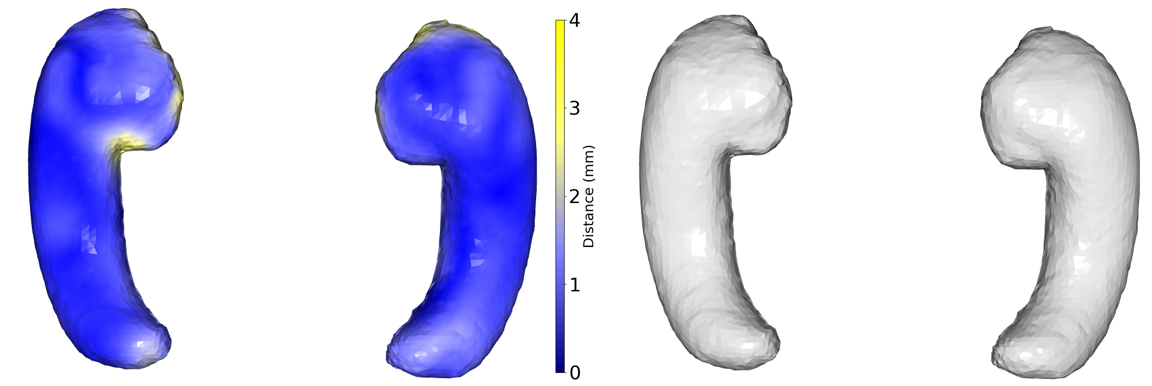}
\caption{On the left side of the figure, we show the combination of reconstructions and original hippocampus (left and right hippocampus) meshes from the 
dataset using our proposed model. The dark blue indicates a very small deviation between the reconstruction and the original mesh. On the right side, we show the original hippocampus data.}
\label{fig:hippo_data}
\end{figure}
\par
The segmentation of the hippocampus in each scan for healthy subjects is conducted automatically \citep{hippo_segment} and manual segmentation is used for the MS subjects, followed by a series of preprocessing steps. Initially, the volumetric representations (i.e., voxel-based) underwent conversion into 3D mesh representations using a marching cubes algorithm \citep{lorensen1998marching_cube}. Subsequently, Laplacian surface smoothing and rigid alignment via an iterative closed point algorithm were applied to eliminate rotational artifacts. To ensure uniform topology across instances, Deformetrica \citep{deformatrica} was employed to establish point correspondence across the meshes due to the assumption of meshes having the same topology \citep{spiralnet++}.
The result of this process is a collection of diffeomorphic deformation maps that illustrate the relationship between a computed mean atlas and the individual subject meshes. Each mesh is characterized by 5944 vertices and 11880 faces. 
In figure \ref{fig:hippo_data}, the color difference illustrates the dissimilarity between original and generated hippocampus shapes, and on the right side, original hippocampus data is shown.

\subsection{Implementation Details}
Our mesh VAE architecture is similar to the SpiralNet++ \citep{spiralnet++}. The encoder module comprises four spiral convolution layers with output channel sizes of [8, 8, 8, 8] and a latent channel size of 12. We use latent channel size as a parameter and find that a size of 12 gives the best results in terms of disentanglement. Sizes smaller than 12 reduce reconstruction accuracy, while sizes larger than 12 compromise disentanglement accuracy. We test sizes of 4, 8, 16, 32, and 12 balanced the trade-off most effectively, providing optimal performance for both disentanglement and reconstruction. The decoder module mirrors the transformations of the encoder. We set $\beta = 0.0015$, according to the parameter tuning results, and employ dilated spiral convolution with subsampling that enhances overall performance. A dilation factor of 2 and a spiral sequence length of 45 are used and those are selected by the parameter tuning process. In the numerical experiments, we adopt an 80/10/10 split for training/validation/testing. The ADAM optimizer is utilized with a batch size of 16, an initial learning rate of $3.6\times10^{-4}$, and a training horizon of 300 epochs. We use a scheduler to decay the primary learning rate by a factor of 0.77 every epoch. The proposed contrastive loss functions use temperature $T = 181$ and threshold $Th=.035$, and all the hyperparameter values are tuned using a hyperparameter optimization framework, Optuna \citep{akiba2019optuna}. We run all the experiments of models on Nvidia Titan RTX GPUs.
\subsection{Evaluation Metrics}
In this section, we present the evaluation metrics used to assess the performance of our proposed model. We focus on disentanglement, regression, and classification aspects, employing a variety of metrics suitable for each task.
\subsubsection{Separated Attribute Predictability (SAP)
\citep{DIPvae}}
SAP score measures how well the model disentangles different attributes or factors of variation. It quantifies the ability to predict individual attributes from the learned representations.
The computation of SAP score involves creating a score matrix, denoted as $S$, of dimensions $R^d \times k$ with $d$ latent variables and $k$ data generative factors. Each entry $(i,j)$ in this matrix signifies the linear regression score for predicting the $j$-th factor using solely the $i$-th latent code. The $R^2$ value of the regression, denoted as $S_{ij}$, represents the predictability. Subsequently, for each column in $S$ (corresponding to a factor), the SAP score is determined as follows:

\begin{equation}
SAP = \frac{1}{M} \sum_{i}^M (S_{i}^* - S_{i}^+)
\end{equation}

In this equation, $S_{i}^*$ is the highest score, $S_{i}^+$ is the second highest, and $M$ denotes the number of considered factors.

\subsubsection{Pearson correlation coefficient (PCC) \citep{pcc}}
We use PCC to calculate the correlation between the values of a specific latent variable and the feature labels (continuous) that the variable is disentangling. The Pearson correlation coefficient, denoted as $r_{xy}$, quantifies the linear relationship between two continuous variables. It measures how well the data points align along a straight line. The formula for Pearson correlation is as follows:
\begin{equation}
    r_{xy} = \frac{\sum_{i=1}^{N} (x_i - \bar{x})(y_i - \bar{y})}{\sqrt{\sum_{i=1}^{N} (x_i - \bar{x})^2 \sum_{i=1}^{N} (y_i - \bar{y})^2}}
\end{equation}
where $N$ is the total number of data points. $x_{i}$ and $y_{i}$ represent the values of the two variables for the $i$-th data point. $\bar{x}$ and $\bar{y}$ denote the means of the $x$ and $y$ values, respectively.

\subsubsection{Point Biserial Correlation (PBC)\citep{pbc}}
PBC is used for the correlation between the values of a specific latent variable and the feature labels (binary) that the variable is disentangling. Point biserial correlation measures the association between a binary attribute and a continuous variable and is defined by the following equation:
\begin{equation}
     r_{pb} = \frac{\bar{X}_1 - \bar{X}_0}{s} \sqrt{\frac{n_1 n_0}{n (n-1)}}
\end{equation}
where $\bar{X}_{1}$ and $\bar{X}_{0}$
are the means of the continuous variable for positive and negative classes, respectively, $s$ is the pooled standard deviation, $n_{1}$ and $n_{0}$ are the sample sizes for positive and negative classes and $n$ is the total sample size.
\subsubsection{Accuracy (Acc.)}
Accuracy measures the proportion of correctly classified instances. We use K-nearest neighbor \citep{knn} for accuracy calculation for our classification task. The values of the specific latent variables are used to predict the labels.

\subsubsection{Mean Squared Error (MSE)}
MSE quantifies the average squared difference between predicted and actual values. The outcomes of K-nearest neighbor \citep{knn} are used for MSE calculation for our regression task. The values of the specific latent variables are used to predict discrete labels.
\begin{equation}
    \text{MSE} = \frac{1}{N} \sum_{i=1}^{N} (y_i - \hat{y}_i)^2
\end{equation}
where $N$ is the total data points and $y_i$ and $\hat{y}_i$ are the original and predicted labels.
\subsubsection{Reconstruction Error (Rec. Err.)}
Rec. Err. measures the dissimilarity (euclidean distance in 3D) between original and reconstructed 3D mesh shapes.
\begin{equation}
\text{Reconstruction Error} = \frac{1}{N} \sum_{i=1}^{N} | x_i - \hat{x}_i |_2^2 
\end{equation}
where $N$ is the total data points and $x_i$ and $\hat{x}_i$ are the original and reconstructed mesh shapes.
\subsubsection{1-Nearest Neighbor Accuracy (1-NNA)\citep{1nna-pointflow}}
1-NNA evaluates the quality of learned representations by comparing nearest neighbors in the learned space using Chamfer Distance (CD) and  Earth Mover’s Distance (EMD). We calculate 1-NNA accuracy using the 3D coordinates from the original data and the 3D coordinates generated from the decoder of our model. We generate the latent variables from our learned distribution of $z$ values from the training set. \par
Let $S_g$ be the set of generated point clouds, and $S_r$ be the set of reference point clouds with $|S_r| = |S_g|$, $S_{-X}$ as the union of $S_r$ and $S_g$ excluding the element $X$, and let $N_X$ represent the nearest neighbor of $X$ within $S_{-X}$. The 1-NN accuracy, denoted as $\text{1-NNA}$, for the 1-NN classifier is expressed as follows:

\begin{equation}
\text{1-NNA}(S_g, S_r) = \frac{\sum_{X \in S_g} I[N_X \in S_g] + \sum_{Y \in S_r} I[N_Y \in S_r]}{|S_g| + |S_r|},
\end{equation}
where $I[\cdot]$ represents the indicator function. In this context, each sample is classified by the 1-NNA classifier as either belonging to $S_r$ or $S_g$ based on the label of its nearest sample. If $S_g$ and $S_r$ are drawn from the same distribution, the accuracy of this classifier should approach 50\% with an adequate number of samples. The proximity of the accuracy to 50\% reflects the similarity between $S_g$ and $S_r$, indicating the model's effectiveness in capturing the target distribution.

CD quantifies the dissimilarity between two point sets by measuring the average distance from each point in one set to its nearest neighbor in the other set.
It is defined as follows:
\begin{equation}
\text{CD}(X, Y) = \frac{1}{|X|} \sum_{x \in X} \min_{y \in Y} |x - y|_2^2 + \frac{1}{|Y|} \sum_{y \in Y} \min_{x \in X} |y - x|_2^2 
\end{equation}
where, $X$ and $Y$ are the two point sets. $|X|$ and $|Y|$ represent the cardinalities of sets $X$ and $Y$, respectively. $|x - y|_2^2$ denotes the Euclidean distance between points $x$ and $y$. \par
EMD, also known as Wasserstein distance, measures the minimum cost required to transform one point distribution into another. It considers the global distribution of points and accounts for both spatial arrangement and quantity.
EMD is defined as: 
\begin{equation}
\text{EMD}(X, Y) = \min_{\gamma} \sum_{(x, y) \in \gamma} c(x, y)
\end{equation}
where, $\gamma$ represents a transport plan that maps points from set $X$ to set $Y$. $c(x, y)$ is the cost of transporting point $x$ to point $y$.

\subsection{Results}
In this section, we present a comprehensive evaluation of our proposed model, Supervised Contrastive VAE (SC VAE), using the evaluation metrics discussed in the previous section. The comparison section compares our method with two baselines and two SOTA methods, using synthetic Torus and Hippocampus (containing both healthy and MS subjects) datasets. The comparison is based on disentanglement, correlation, prediction, reconstruction, and data-generative performance. The baseline models are $\beta$-VAE \citep{betavae} and $\beta$-TCVAE \citep{betatcvae} while we use Supervised Guided VAE (SG VAE) \citep{guidedvae} and Attribute VAE \citep{attributevae} as SOTA methods.
\par
We provide an ablation study of our method, demonstrating the significance of the inhibition term. Furthermore, we present individual SAP scores for the discrete and continuous labels, when our model is trained to disentangle them separately. Then the training and test time comparisons are shown for all the five models. Lastly, we demonstrate the implementation of our model to analyze 3D hippocampus shape changes due to MS and aging.
\subsubsection{Comparison}
We compare the models in terms of disentanglement score (SAP), the correlation between the latent variables and labels (Corr.), accuracy (Acc.), and MSE score in predicting the labels from the values of the latent variables using the K-nearest neighbor classifier. The reported SAP scores in table \ref{table_disentangle_corr_acc} are calculated by averaging the SAP scores for $z_1$ and $z_2$ variables and the models are trained simultaneously for classification and regression tasks for all the scores. \par

The results, presented in Table \ref{table_disentangle_corr_acc}, include SAP score (average of classification and regression SAP scores), correlation, accuracy, and MSE for all five models across both torus and hippocampus datasets. Our model demonstrates superior performance in SAP scores for both datasets while achieving comparable or better results in terms of correlation, accuracy, and MSE compared to the other models. 
\begin{table}[h!]
    \centering
    \caption{Comparison among models using SAP scores, correlation, accuracy, and MSE utilizing the hippocampus and synthetic torus dataset. A higher score is better ($\uparrow$) for all the metrics except MSE (a lower score is better for MSE ($\downarrow$)).}
    \vspace{3mm}
    \begin{tabular}{cccccc}
    \hline
    \\[-1ex]
    \vspace{1mm}
        \textbf{Model} & 
        \textbf{Dataset} &
        \textbf{SAP} & 
        \textbf{Corr.} &
        \textbf{Acc.} &
        \textbf{MSE}
        \\
         & & ($\uparrow$) & ($\uparrow$) & ($\uparrow$) & ($\downarrow$) \\
        
        \\[-1ex]
        \hline
        \\[-1ex]
        \multirow{2}{*}{\textbf{$\mathbf{\beta}$-VAE}} & Torus  & 0.43 & 0.48 &  64.47 & 0.074 \\ 
        & Hippocampus  & 0.09 & 0.38 &  53.98 & 0.091 \\
        \\[-1ex]
        \hline
        \\[-1ex]
        \multirow{2}{*}{\textbf{$\mathbf{\beta}$-TCVAE}} & Torus  & 0.45 & 0.48 &  71.95 & 0.071\\ 
        & Hippocampus  & 0.11 & 0.39 &  53.23 & 0.093\\
        \\[-1ex]
        \hline
        \\[-1ex]
        \multirow{2}{*}{\textbf{SG VAE}} & Torus  & 0.64 & \textbf{0.78} &  \textbf{100} & \textbf{0.013} \\ 
        & Hippocampus  & 0.31 & 0.69 &  98.09 & 0.029  \\
        \\[-1ex]
        \hline
        \\[-1ex]
        \multirow{2}{*}{\textbf{Attribute VAE}} & Torus &  0.66 & 0.74 &  \textbf{100} & 0.017 \\
        & Hippocampus  & 0.32 & 0.66 &  98.13 & 0.028 \\
        \\[-1ex]
        \hline
        \\[-1ex]
        \multirow{2}{*}{\textbf{SC VAE (Ours)}} & Torus  & \textbf{0.69} & 0.75 & \textbf{100} & 0.016 \\
        & Hippocampus & \textbf{0.36}   & \textbf{0.70} & \textbf{98.31}   & \textbf{0.025} \\
        \\[-1ex]
        \hline
    \end{tabular}
    \label{table_disentangle_corr_acc}
\end{table}
\par
In Table \ref{rec_comparison}, we present the reconstruction error and 1-NNA scores using CD and EMD for all five models across both datasets. Lower values indicate better performance for both reconstruction and 1-NNA scores. Our model demonstrates superior 1-NNA scores for the torus dataset using EMD. For both datasets, our model's scores, except for 1-NNA on the torus dataset, are either better or comparable to those of supervised models. However, baseline models exhibit better performance in terms of reconstruction error and the quality of data generation (1-NNA). These results align with expectations, as increased disentanglement in the latent space poses a challenge for models to simultaneously reduce reconstruction error and maintain high-quality data generation capabilities. \par

To summarize, our proposed method performs better than all four methods in terms of disentanglement, while also showing comparable or better results in terms of prediction and correlation. Additionally, our method performs similarly in terms of reconstruction error and data generation quality when compared to supervised disentangled methods.

\begin{table}[h!]
    \centering
    \caption{Comparison of reconstruction and generative quality among models using both of our datasets. Reconstruction Error (Rec. Err.) and 1-NNA scores using CD and EMD are calculated for every model. Lower scores are better ($\downarrow$) for each model.}
    \vspace{3mm}
    \begin{tabular}{ccccc}
    \hline
    \\[-1ex]
    \vspace{1mm}
        \textbf{Model} & 
        \textbf{Dataset} &
        \textbf{Rec. Err.} &
        \multicolumn{2}{c}{\textbf{1-NNA(\%, $\downarrow$)}} 
        \\
         & &  ($\downarrow$) & \textbf{CD} & \textbf{EMD} \\
        
        \\[-1ex]
        \hline
        \\[-1ex]
        \multirow{2}{*}{\textbf{$\mathbf{\beta}$-VAE}} & Torus   &  \textbf{0.25} & \textbf{51.37} & 56.25\\ 
        & Hippocampus   &  \textbf{0.85} & 56.58 & 55.96 \\
        \\[-1ex]
        \hline
        \\[-1ex]
        \multirow{2}{*}{\textbf{$\mathbf{\beta}$-TCVAE}} & Torus   &  0.28 & 52.35 & 54.71\\ 
        & Hippocampus   &  0.86 & \textbf{56.35} & \textbf{55.43}\\
        \\[-1ex]
        \hline
        \\[-1ex]
        \multirow{2}{*}{\textbf{SG VAE}} & Torus  &  0.33 & 52.78 & 56.33\\ 
        & Hippocampus   &  1.08 & 61.79 & 60.05 \\
        \\[-1ex]
        \hline
        \\[-1ex]
        \multirow{2}{*}{\textbf{Attribute VAE}} & Torus & 0.36 & 59.38 & 57.81 \\
        & Hippocampus   &  1.09 & 59.38 & 58.37 \\
        \\[-1ex]
        \hline
        \\[-1ex]
        \multirow{2}{*}{\textbf{SC VAE (Ours)}} & Torus   & 0.38 & 51.56 & \textbf{53.12} \\
        & Hippocampus  & 1.07 & 58.69 & 59.76\\
        \\[-1ex]
        \hline
    \end{tabular}
    \label{rec_comparison}
\end{table}

\subsubsection{Ablation Study}
Our ablation study shows the significance of the inhibition term using different metrics. In Table \ref{ablation_study}, we show the ablation study, providing the scores of SAP, correlation, accuracy, MSE, and reconstruction errors for both torus and hippocampus datasets. Introducing the additional denominator term (inhibition when $\lambda_{1}=1$ and $\lambda_{2}=1$ in equations \ref{eq:9} and \ref{eq:10}) results in an increase in SAP score compared to using the SNN loss without any modification (w/o inhibition when $\lambda_{1}=1$ and $\lambda_{2}=0$ in equations \ref{eq:9} and \ref{eq:10}) for both datasets. Meanwhile, the scores of other metrics remain comparable. Here, lower MSE and reconstruction error scores signify improved performance. \par
\begin{figure}[h!]
    \centering
    \includegraphics[width=\linewidth]{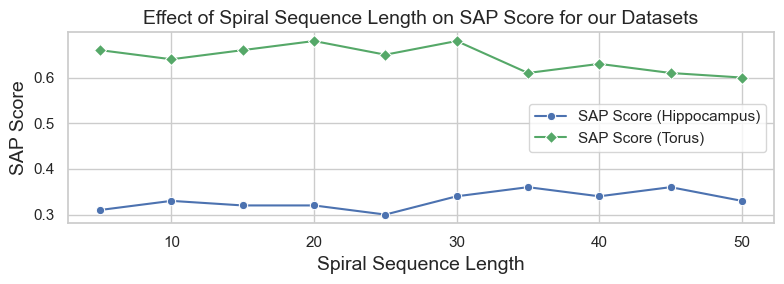}
    \caption{Effect of Spiral Sequence Length on SAP Score for our Torus and Hippocampus Datasets.
    }
\label{fig:SAP_score_seq_length}
\end{figure}
Additionally, we conducted an ablation study to verify the effect of neighborhood information on disentanglement in mesh-based convolutional neural networks. The plot in figure \ref{fig:SAP_score_seq_length} demonstrates that varying the spiral sequence length affects the SAP score for both the hippocampus and torus datasets. The non-linear relationship between spiral length and SAP scores shows that optimal performance occurs at specific lengths, implying that neighborhood connectivity impacts disentanglement. \textcolor{blue}{Therefore, the length of the spiral, as neighborhood information, has a significant effect on disentanglement. On a scale of 1, the SAP score varies in the range of 0.07 and 0.06 for the torus and hippocampus data, respectively. The scores would not change much or remain similar if a reasonable spiral length does not affect disentanglement.} This justifies our use of SpiralNet++, which leverages such connectivity effectively.
\begin{table}[h!]
    \centering
    \caption{Ablation Study of SC VAE by including and excluding the inhibition term in the denominator of our loss function. Higher scores are better ($\uparrow$) for SAP, Corr. and Acc. and lower scores are better ($\downarrow$) for MSE and Rec. Err. }
    \vspace{3mm}
    \begin{tabular}{ccccccc}
    \\[-1ex]
    \hline
    \\[-1ex]
    \vspace{1mm}
        \textbf{Model} & 
        \textbf{Dataset} &
        \textbf{SAP} & 
        \textbf{Corr.} &
        \textbf{Acc.} &
        \textbf{MSE} &
        \textbf{Rec. Err.}\\

        & &  ($\uparrow$) &  ($\uparrow$)&  ($\uparrow$)&  ($\downarrow$)&  ($\downarrow$) \\
                \\[-1ex]
        \hline
        \\[-1ex]
        \multirow{2}{*}{\textbf{SC VAE ($\mathbf{w/o \hspace{1mm} inhibition}$)}} & Torus & 0.66  & \textbf{0.77} & \textbf{100} & \textbf{0.016} & \textbf{0.37} \\
        & Hippocampus  & 0.31 & \textbf{0.70} & 98.11 & 0.027 & \textbf{1.07} \\
        \\[-1ex]
        \hline
        \\[-1ex]
        \multirow{2}{*}{\textbf{SC VAE (w/ inhibition)}}  & Torus & \textbf{0.68}  & 0.75 & \textbf{100}  & \textbf{0.016} &0.38 \\
        & Hippocampus  & \textbf{0.36} & \textbf{0.70} & \textbf{98.31} & \textbf{0.025} & \textbf{1.07} \\
        \\[-1ex]
        \hline
    \end{tabular}
    \label{ablation_study}
\end{table}

\subsubsection{Training and Test Time}
We present the training (seconds per epoch) and test times (seconds per test set) for both the synthetic torus (80\% of 5000 instances during training, and 10\% each for test and validation) and hippocampus datasets (80\% of 553 instances during training, and 10\% each for test and validation). Guided VAE exhibits the longest training time due to additional parameters introduced by neural networks for excitation and inhibition mechanisms. Our method requires slightly more time than Attribute VAE during training, while the two baseline methods outperform supervised methods in terms of training time. During test time, our method demonstrates performance comparable to other methods.
\begin{table}[h!]
    \centering
    \caption{Training and Test Time (in seconds) for Torus and Hippocampus Dataset. A lower score is better for all the models.}
    \vspace{3mm}
    \begin{tabular}{cccccccc}
    \\[-1ex]
    \hline
    \\[-1ex]
    \vspace{1mm}
        \multirow{2}{*}{\textbf{Models}} & &
        $\beta$-VAE &
        $\beta$-TC- & 
         SG &
        Attribute &
        SC \\
        & & & VAE & VAE & VAE&VAE \\
        & & &  &  & &(Ours) \\
        \\[-1ex]
        \hline
        \\[-1ex]
        \multirow{2}{*}{\textbf{Training (Sec./Epoch)}} & Torus & \textbf{6.93} & 7.89 & 23.14 & 9.8 & 10.50\\
        & Hippocampus & \textbf{0.28} & 0.29 & 0.91 & 0.33 & 0.49\\
        \\[-1ex]
        \hline
        \\[-1ex]
         \multirow{2}{*}{\textbf{Test (Sec./Set)}} & Torus & \textbf{0.24} & \textbf{0.24} & 0.26 & 0.25 & 0.25 \\
       & Hippocampus  & \textbf{0.02} & \textbf{0.02} & 0.03 & 0.03 & 0.03 \\
        \\[-1ex]
        \hline
    \end{tabular}
    \label{train_test_time}
\end{table}
\subsubsection{SAP Scores for Classification and Regression Tasks}
\begin{figure}[h!]
    \centering
    \includegraphics[width=\linewidth]{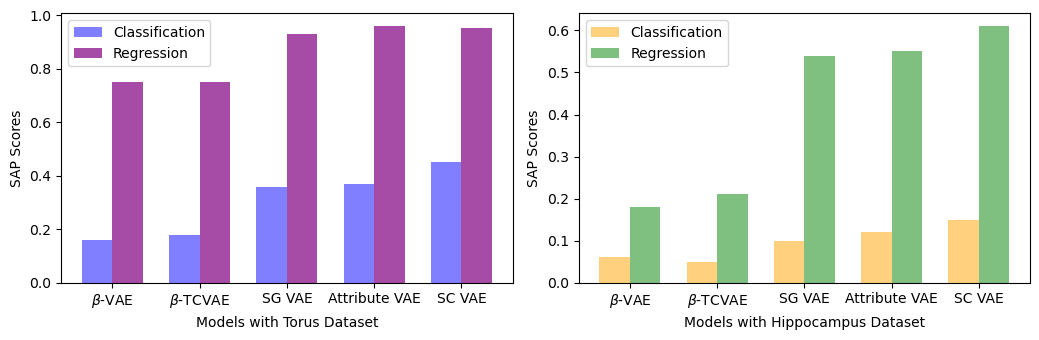}
    \caption{SAP scores (Classification and regression are separated) for different models using synthetic torus (left) and hippocampus (right) datasets.
    }
\label{fig:SAP_score_separated}
\end{figure}

We present the mean SAP score within the comparison section. Figure \ref{fig:SAP_score_separated} presents the classification and regression scores for both torus and hippocampus datasets when trained independently (All the other results in different sections were obtained by training classification and regression tasks simultaneously). Our approach demonstrates superior performance in classification and regression tasks for both datasets compared to other methods, except for Attribute VAE, which does a slightly better job in the regression task specifically for the Torus dataset.

\subsubsection{MS vs Hippocampal Volume Across Age Groups}
\begin{figure}[h!]
    \centering
    \includegraphics[width=\linewidth]{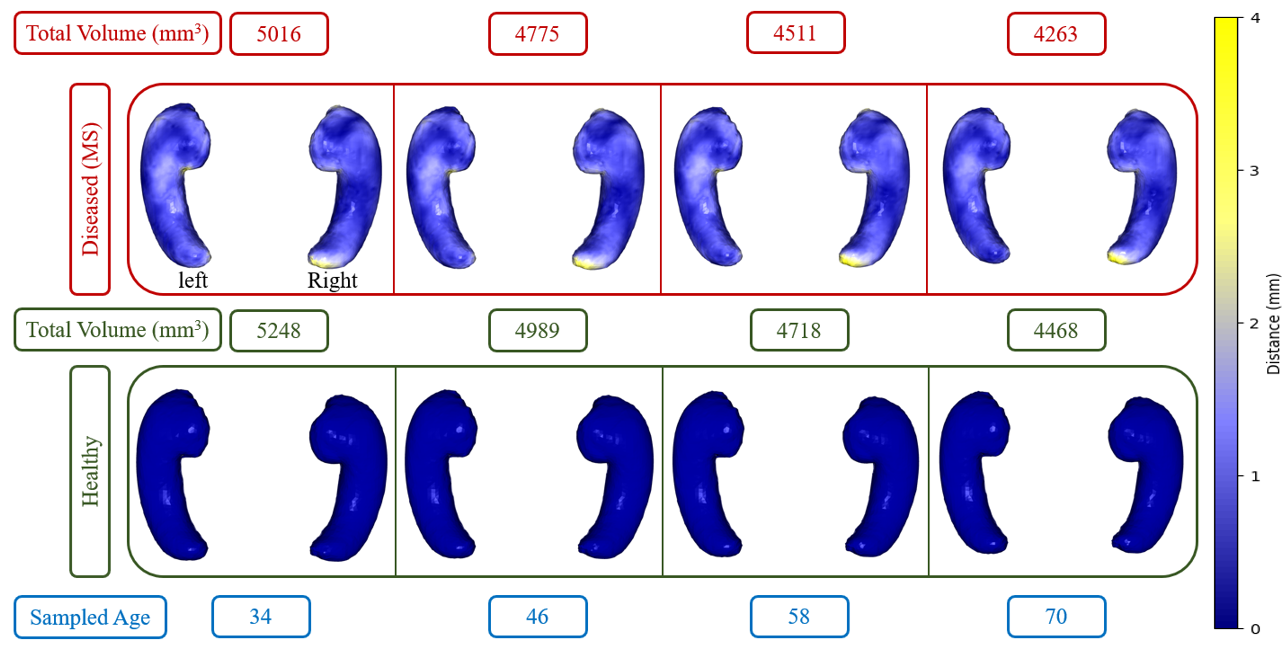}
    \caption{Volume changes (between healthy and MS) are depicted in the first row by the intensity of the blue color and yellow represents the highest change in millimeters. The second row shows the healthy hippocampus. Ages are calculated by mapping the latent values and age range of the subjects of MS.}
    \label{fig:MS_age}
\end{figure}
We employ our trained model to produce hippocampus shapes and assess the data-generating capabilities of our model within the domain of medical data. Our trained model demonstrates the ability to capture volume changes according to different data generative factors like age and diseases. By mapping the latent variable $z_2$ values within the range of -3$\sigma$ to +3$\sigma$ onto the age range of MS subjects, we select ages with intervals. Subsequently, we obtain the shapes of the healthy hippocampus by fixing the $z_1$ value at -3$\sigma$ and the MS hippocampus by maintaining the $z_1$ value at 3$\sigma$.\par
\begin{figure}[h!]
    \centering
    \includegraphics[width=0.99\linewidth]{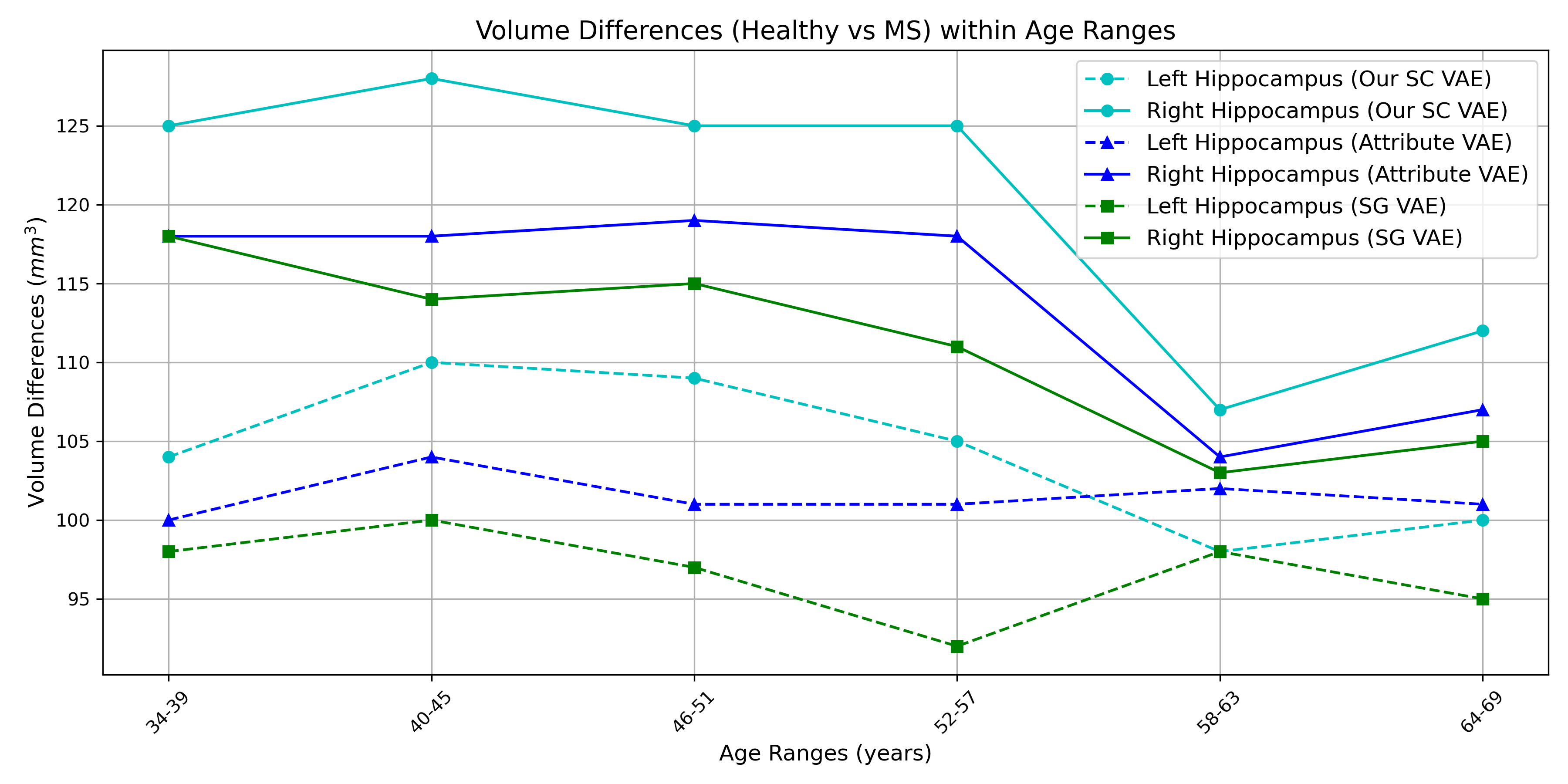}
    \caption{Volume differences between healthy and MS hippocampus are shown using six age ranges. Volume differences for the left and right hippocampus are shown separately for all three supervised disentanglement methods (SG VAE, Attribute VAE, and our SC VAE). The plot depicts a higher right hippocampus volume difference across ages derived by all three methods. The volumes are calculated by generating 10 shapes (using the three models) from each range, and then taking the average. Our method shows the highest volume difference \textcolor{blue}{($\sim$4.5\%)} among the methods and is closer (compared to other methods) to the average trend \textcolor{blue}{(9\%)} found in a study by \citet{MS_hippocampal_change} that calculated hippocampus volumes from DTI.}
    \label{fig:MS_tle_volume}
\end{figure}

In Figure \ref{fig:MS_age}, we present the results for MS vs healthy controls. The lower row displays healthy hippocampus shapes for four sample ages, while the upper row depicts MS hippocampus shapes at those ages. Volume changes (between healthy and MS) are depicted in the first row by the intensity of the blue color and yellow represents the highest change in millimeters. The figure illustrates a noticeable decrease in hippocampus size with advancing age, particularly affecting the right hippocampus in MS. The tail of the right hippocampus has more atrophy than other regions. \par Our findings align with previous research by \citet{MS_right_reason} and \citet{mslefthippo}, which reported a larger reduction in the volume of the right hippocampus compared to the left hippocampus due to MS. Additionally, the overall hippocampus volume is lower in the MS population according to our results and the reduction of volume is also reported in the findings of \citet{MS_hippocampal_change}. However, the volume reduction \textcolor{blue}{($\sim$4.5\%)} by our method is lower than \citet{MS_hippocampal_change} \textcolor{blue}{(9\%)} who used DTI for calculating volumes, and it is expected because our classification (presence or absence of MS) SAP score is lower than the regression (age) SAP score.
Therefore, we need more data for the MS population for more accurate results. We also show volume differences in Figure \ref{fig:MS_tle_volume} for the three supervised disentangled methods (SG VAE, Attribute VAE, and our SC VAE) and our method shows the highest volume difference, which is closer (compared to other methods) to the average trend found in \citep{MS_hippocampal_change}. Additionally, we conducted a one-sample t-test \citep{1-samplettest} to examine whether there was a significant difference in hippocampal volume between healthy and MS populations, and we found the differences are significant ($p<.001$). In Figure \ref{fig:MS_tle_volume}, the plot illustrates the greater decrease in the volume of the right hippocampus compared to the left hippocampus in patients with MS. We generate the plot by generating 10 shapes (using the three models) from each range, calculating the volume and then take the average volume. \par

\section{Discussion and Conclusion}
\subsection{Discussion}
Our study contributes to the field of medical imaging and shape analysis in several ways. The proposed method enhances disentanglement performance for categorical and continuous labels in the context of 3D mesh data. The analysis of anatomical shape variations across various factors, including age and disease (MS), through the generation of 3D shapes, provides valuable insights into the relationship between neurological disorders and hippocampal shape changes.
\par 
Despite the promising results, there are limitations to our approach. The generalization of our model to diverse populations and datasets needs further exploration. We also need to improve the disentanglement performance of disease status for better prediction and reconstruction. Additionally, the mesh convolution technique we used, as outlined by \citet{spiralnet++}, necessitates the registration of meshes to a template mesh. Consequently, it is crucial to investigate methods that do not rely on assumptions about mesh topology for the analysis of complex shapes. \textcolor{blue}{Also, Our classification and regression losses are mostly similar, with only minor differences. Therefore, a unified formulation could be utilized to compactly represent the loss function.} Future work would involve the incorporation of longitudinal data and exploration of the generalizability of the proposed method.
\subsection{Conclusion}
In this paper, we propose a novel approach for disentangling 3D mesh shape (hippocampal or synthetic) variations from DTI or synthetic datasets and applied in the context of neurological disorders. Our method, which uses a Mesh VAE enhanced with Supervised Contrastive Learning, exhibits superior disentanglement capabilities, particularly in identifying age and disease status in patients with MS. Additionally, our method demonstrates comparable or better performance in all other metrics. The validity of our method is also demonstrated by a synthetic torus dataset. 
\par 
 We aim to extract meaningful representations of anatomical structures, providing insights into the complexities of diseases and age-related variations in the hippocampus by integrating novel and efficient latent space disentanglement techniques. Our method demonstrates the extraction of valuable insights into hippocampal morphology and atrophy linked to age and MS, even in the face of challenges posed by the absence of longitudinal data in limited datasets.

\acks{Operating grant was provided by the Canadian Institutes of Health Research (CIHR) and the DTI dataset acquisition was funded by the University Hospital Foundation and the Women and Children’s Health Research Institute. Author DC acknowledges student funding from Natural Sciences and Engineering Research Council DG program. Author CB acknowledges funding by CIHR and Canada Research Chairs. Infrastructure was provided by the Canada Foundation for Innovation, Alberta Innovation and Advanced Education, and the University Hospital Foundation. }

%
\ethics{No ethics approval was required for the synthetic data analysis. The neuroimaging study was approved by the Human
Research Ethics Board at the University of Alberta}

\coi{We declare we don’t have conflicts of interest.}

\data{The GitHub repository contains the script for generating synthetic data. Although the hippocampus data is confidential and cannot be shared, the preprocessing scripts are provided and can be utilized for any publicly available MRI data that includes hippocampus segmentation.}

\bibliography{sample}

\begin{thebibliography}{39}
\providecommand{\natexlab}[1]{#1}
\providecommand{\url}[1]{\texttt{#1}}
\expandafter\ifx\csname urlstyle\endcsname\relax
  \providecommand{\doi}[1]{doi: #1}\else
  \providecommand{\doi}{doi: \begingroup \urlstyle{rm}\Url}\fi

\bibitem[Akiba et~al.(2019)Akiba, Sano, Yanase, Ohta, and Koyama]{akiba2019optuna}
Takuya Akiba, Shotaro Sano, Toshihiko Yanase, Takeru Ohta, and Masanori Koyama.
\newblock Optuna: A next-generation hyperparameter optimization framework.
\newblock In \emph{Proceedings of the 25th ACM SIGKDD international conference on knowledge discovery \& data mining}, pages 2623--2631, 2019.

\bibitem[Altaf et~al.(2019)Altaf, Islam, Akhtar, and Janjua]{intro_shape_analysis}
Fouzia Altaf, Syed~MS Islam, Naveed Akhtar, and Naeem~Khalid Janjua.
\newblock Going deep in medical image analysis: concepts, methods, challenges, and future directions.
\newblock \emph{IEEE Access}, 7:\penalty0 99540--99572, 2019.

\bibitem[Aneja et~al.(2021)Aneja, Schwing, Kautz, and Vahdat]{contrastive2}
Jyoti Aneja, Alex Schwing, Jan Kautz, and Arash Vahdat.
\newblock A contrastive learning approach for training variational autoencoder priors.
\newblock \emph{Advances in neural information processing systems}, 34:\penalty0 480--493, 2021.

\bibitem[Brown(2001)]{pbc}
James~Dean Brown.
\newblock Point-biserial correlation coefficients.
\newblock \emph{Statistics}, 5\penalty0 (3):\penalty0 12--6, 2001.

\bibitem[Burgess et~al.(2018)Burgess, Higgins, Pal, Matthey, Watters, Desjardins, and Lerchner]{annealed-vae}
Christopher~P Burgess, Irina Higgins, Arka Pal, Loic Matthey, Nick Watters, Guillaume Desjardins, and Alexander Lerchner.
\newblock Understanding disentangling in $beta $-vae.
\newblock \emph{arXiv preprint arXiv:1804.03599}, 2018.

\bibitem[Cetin et~al.(2023)Cetin, Stephens, Camara, and Ballester]{attributevae}
Irem Cetin, Maialen Stephens, Oscar Camara, and Miguel A~Gonz{\'a}lez Ballester.
\newblock Attri-vae: Attribute-based interpretable representations of medical images with variational autoencoders.
\newblock \emph{Computerized Medical Imaging and Graphics}, 104:\penalty0 102158, 2023.

\bibitem[Chen et~al.(2018)Chen, Li, Grosse, and Duvenaud]{betatcvae}
Ricky~TQ Chen, Xuechen Li, Roger~B Grosse, and David~K Duvenaud.
\newblock Isolating sources of disentanglement in variational autoencoders.
\newblock \emph{Advances in neural information processing systems}, 31, 2018.

\bibitem[Chorin et~al.(2020)Chorin, Dai, Shulman, Wadhwani, Bar-Cohen, Barbhaiya, Aizer, Holmes, Bernstein, Spinelli, et~al.]{1-samplettest}
Ehud Chorin, Matthew Dai, Eric Shulman, Lalit Wadhwani, Roi Bar-Cohen, Chirag Barbhaiya, Anthony Aizer, Douglas Holmes, Scott Bernstein, Michael Spinelli, et~al.
\newblock The qt interval in patients with covid-19 treated with hydroxychloroquine and azithromycin.
\newblock \emph{Nature medicine}, 26\penalty0 (6):\penalty0 808--809, 2020.

\bibitem[Cohen et~al.(2009)Cohen, Huang, Chen, Benesty, Benesty, Chen, Huang, and Cohen]{pcc}
Israel Cohen, Yiteng Huang, Jingdong Chen, Jacob Benesty, Jacob Benesty, Jingdong Chen, Yiteng Huang, and Israel Cohen.
\newblock Pearson correlation coefficient.
\newblock \emph{Noise reduction in speech processing}, pages 1--4, 2009.

\bibitem[Deng et~al.(2020)Deng, Yang, Chen, Wen, and Tong]{contrastive1}
Yu~Deng, Jiaolong Yang, Dong Chen, Fang Wen, and Xin Tong.
\newblock Disentangled and controllable face image generation via 3d imitative-contrastive learning.
\newblock In \emph{Proceedings of the IEEE/CVF conference on computer vision and pattern recognition}, pages 5154--5163, 2020.

\bibitem[Ding et~al.(2020)Ding, Xu, Xu, Parmar, Yang, Welling, and Tu]{guidedvae}
Zheng Ding, Yifan Xu, Weijian Xu, Gaurav Parmar, Yang Yang, Max Welling, and Zhuowen Tu.
\newblock Guided variational autoencoder for disentanglement learning.
\newblock In \emph{Proceedings of the IEEE/CVF conference on computer vision and pattern recognition}, pages 7920--7929, 2020.

\bibitem[Dupont(2018)]{jointvae}
Emilien Dupont.
\newblock Learning disentangled joint continuous and discrete representations.
\newblock \emph{Advances in neural information processing systems}, 31, 2018.

\bibitem[Durrleman et~al.(2014)Durrleman, Prastawa, Charon, Korenberg, Joshi, Gerig, and Trouv{\'e}]{deformatrica}
Stanley Durrleman, Marcel Prastawa, Nicolas Charon, Julie~R Korenberg, Sarang Joshi, Guido Gerig, and Alain Trouv{\'e}.
\newblock Morphometry of anatomical shape complexes with dense deformations and sparse parameters.
\newblock \emph{NeuroImage}, 101:\penalty0 35--49, 2014.

\bibitem[Efird et~al.(2021)Efird, Neumann, Solar, Beaulieu, and Cobzas]{hippo_segment}
Cory Efird, Samuel Neumann, Kevin~G Solar, Christian Beaulieu, and Dana Cobzas.
\newblock Hippocampus segmentation on high resolution diffusion mri.
\newblock In \emph{2021 IEEE 18th International Symposium on Biomedical Imaging (ISBI)}, pages 1369--1372. IEEE, 2021.

\bibitem[Estermann and Wattenhofer(2023)]{dava}
Benjamin Estermann and Roger Wattenhofer.
\newblock Dava: Disentangling adversarial variational autoencoder.
\newblock \emph{arXiv preprint arXiv:2303.01384}, 2023.

\bibitem[Foti et~al.(2022)Foti, Koo, Stoyanov, and Clarkson]{3Dvaeswap}
Simone Foti, Bongjin Koo, Danail Stoyanov, and Matthew~J Clarkson.
\newblock 3d shape variational autoencoder latent disentanglement via mini-batch feature swapping for bodies and faces.
\newblock In \emph{Proceedings of the IEEE/CVF conference on computer vision and pattern recognition}, pages 18730--18739, 2022.

\bibitem[Frosst et~al.(2019)Frosst, Papernot, and Hinton]{SNN_contrastive_loss}
Nicholas Frosst, Nicolas Papernot, and Geoffrey Hinton.
\newblock Analyzing and improving representations with the soft nearest neighbor loss.
\newblock In \emph{International conference on machine learning}, pages 2012--2020. PMLR, 2019.

\bibitem[Gong et~al.(2019)Gong, Chen, Bronstein, and Zafeiriou]{spiralnet++}
Shunwang Gong, Lei Chen, Michael Bronstein, and Stefanos Zafeiriou.
\newblock Spiralnet++: A fast and highly efficient mesh convolution operator.
\newblock In \emph{Proceedings of the IEEE/CVF international conference on computer vision workshops}, pages 0--0, 2019.

\bibitem[Higgins et~al.(2016)Higgins, Matthey, Pal, Burgess, Glorot, Botvinick, Mohamed, and Lerchner]{betavae}
Irina Higgins, Loic Matthey, Arka Pal, Christopher Burgess, Xavier Glorot, Matthew Botvinick, Shakir Mohamed, and Alexander Lerchner.
\newblock beta-vae: Learning basic visual concepts with a constrained variational framework.
\newblock In \emph{International conference on learning representations}, 2016.

\bibitem[Huang et~al.(2023)Huang, Jin, Lu, Hou, Cheng, Fu, Shen, and Feng]{contrastive3}
Zhicheng Huang, Xiaojie Jin, Chengze Lu, Qibin Hou, Ming-Ming Cheng, Dongmei Fu, Xiaohui Shen, and Jiashi Feng.
\newblock Contrastive masked autoencoders are stronger vision learners.
\newblock \emph{IEEE Transactions on Pattern Analysis and Machine Intelligence}, 2023.

\bibitem[Hulst et~al.(2015)Hulst, Schoonheim, Van~Geest, Uitdehaag, Barkhof, and Geurts]{mslefthippo}
Hanneke~E Hulst, Menno~M Schoonheim, Quinten Van~Geest, Bernard~MJ Uitdehaag, Frederik Barkhof, and Jeroen~JG Geurts.
\newblock Memory impairment in multiple sclerosis: relevance of hippocampal activation and hippocampal connectivity.
\newblock \emph{Multiple Sclerosis Journal}, 21\penalty0 (13):\penalty0 1705--1712, 2015.

\bibitem[Imandoust et~al.(2013)Imandoust, Bolandraftar, et~al.]{knn}
Sadegh~Bafandeh Imandoust, Mohammad Bolandraftar, et~al.
\newblock Application of k-nearest neighbor (knn) approach for predicting economic events: Theoretical background.
\newblock \emph{International journal of engineering research and applications}, 3\penalty0 (5):\penalty0 605--610, 2013.

\bibitem[Kiechle et~al.(2023)Kiechle, Miller, Slessor, Pietrosanu, Kong, Beaulieu, and Cobzas]{intro_shape_analysis_age}
Johannes Kiechle, Dylan Miller, Jordan Slessor, Matthew Pietrosanu, Linglong Kong, Christian Beaulieu, and Dana Cobzas.
\newblock Explaining anatomical shape variability: Supervised disentangling with a variational graph autoencoder.
\newblock In \emph{2023 IEEE 20th International Symposium on Biomedical Imaging (ISBI)}, pages 1--5. IEEE, 2023.

\bibitem[Kim and Mnih(2018)]{factorvae}
Hyunjik Kim and Andriy Mnih.
\newblock Disentangling by factorising.
\newblock In \emph{International Conference on Machine Learning}, pages 2649--2658. PMLR, 2018.

\bibitem[Kingma et~al.(2019)Kingma, Welling, et~al.]{vae}
Diederik~P Kingma, Max Welling, et~al.
\newblock An introduction to variational autoencoders.
\newblock \emph{Foundations and Trends{\textregistered} in Machine Learning}, 12\penalty0 (4):\penalty0 307--392, 2019.

\bibitem[Kipf and Welling(2016)]{vgae}
Thomas~N Kipf and Max Welling.
\newblock Variational graph auto-encoders.
\newblock \emph{arXiv preprint arXiv:1611.07308}, 2016.

\bibitem[Kumar et~al.(2017)Kumar, Sattigeri, and Balakrishnan]{DIPvae}
Abhishek Kumar, Prasanna Sattigeri, and Avinash Balakrishnan.
\newblock Variational inference of disentangled latent concepts from unlabeled observations.
\newblock \emph{arXiv preprint arXiv:1711.00848}, 2017.

\bibitem[Litany et~al.(2018)Litany, Bronstein, Bronstein, and Makadia]{litany2018deformable}
Or~Litany, Alex Bronstein, Michael Bronstein, and Ameesh Makadia.
\newblock Deformable shape completion with graph convolutional autoencoders.
\newblock In \emph{Proceedings of the IEEE conference on computer vision and pattern recognition}, pages 1886--1895, 2018.

\bibitem[Lorensen and Cline(1998)]{lorensen1998marching_cube}
William~E Lorensen and Harvey~E Cline.
\newblock Marching cubes: A high resolution 3d surface construction algorithm.
\newblock In \emph{Seminal graphics: pioneering efforts that shaped the field}, pages 347--353. 1998.

\bibitem[Lv et~al.(2021)Lv, Lin, and Zhao]{mesh_surface_detail}
Chenlei Lv, Weisi Lin, and Baoquan Zhao.
\newblock Voxel structure-based mesh reconstruction from a 3d point cloud.
\newblock \emph{IEEE Transactions on Multimedia}, 24:\penalty0 1815--1829, 2021.

\bibitem[Oord et~al.(2018)Oord, Li, and Vinyals]{infonce}
Aaron van~den Oord, Yazhe Li, and Oriol Vinyals.
\newblock Representation learning with contrastive predictive coding.
\newblock \emph{arXiv preprint arXiv:1807.03748}, 2018.

\bibitem[Pan et~al.(2018)Pan, Hu, Long, Jiang, Yao, and Zhang]{vgae2}
Shirui Pan, Ruiqi Hu, Guodong Long, Jing Jiang, Lina Yao, and Chengqi Zhang.
\newblock Adversarially regularized graph autoencoder for graph embedding.
\newblock \emph{arXiv preprint arXiv:1802.04407}, 2018.

\bibitem[Roosendaal et~al.(2010)Roosendaal, Hulst, Vrenken, Feenstra, Castelijns, Pouwels, Barkhof, and Geurts]{MS_right_reason}
Stefan~D Roosendaal, Hanneke~E Hulst, Hugo Vrenken, Heleen~EM Feenstra, Jonas~A Castelijns, Petra~JW Pouwels, Frederik Barkhof, and Jeroen~JG Geurts.
\newblock Structural and functional hippocampal changes in multiple sclerosis patients with intact memory function.
\newblock \emph{Radiology}, 255\penalty0 (2):\penalty0 595--604, 2010.

\bibitem[Solar et~al.(2021)Solar, Treit, and Beaulieu]{hippocampus_data_acquisition}
Kevin~Grant Solar, Sarah Treit, and Christian Beaulieu.
\newblock High resolution diffusion tensor imaging of the hippocampus across the healthy lifespan.
\newblock \emph{Hippocampus}, 31\penalty0 (12):\penalty0 1271--1284, 2021.

\bibitem[Sun et~al.(2022)Sun, Pears, and Gu]{3dvaesecondface}
Hao Sun, Nick Pears, and Yajie Gu.
\newblock Information bottlenecked variational autoencoder for disentangled 3d facial expression modelling.
\newblock In \emph{Proceedings of the IEEE/CVF Winter Conference on Applications of Computer Vision}, pages 157--166, 2022.

\bibitem[Vald{\'e}s~Cabrera et~al.(2023)Vald{\'e}s~Cabrera, Blevins, Smyth, Emery, Solar, and Beaulieu]{MS_hippocampal_change}
Diana Vald{\'e}s~Cabrera, Gregg Blevins, Penelope Smyth, Derek Emery, Kevin~Grant Solar, and Christian Beaulieu.
\newblock High-resolution diffusion tensor imaging and t2 mapping detect regional changes within the hippocampus in multiple sclerosis.
\newblock \emph{NMR in Biomedicine}, 36\penalty0 (9):\penalty0 e4952, 2023.

\bibitem[Van~der Velden et~al.(2022)Van~der Velden, Kuijf, Gilhuijs, and Viergever]{intro_shape_analysis_dis}
Bas~HM Van~der Velden, Hugo~J Kuijf, Kenneth~GA Gilhuijs, and Max~A Viergever.
\newblock Explainable artificial intelligence (xai) in deep learning-based medical image analysis.
\newblock \emph{Medical Image Analysis}, 79:\penalty0 102470, 2022.

\bibitem[Wang et~al.(2017)Wang, Pan, Long, Zhu, and Jiang]{wang2017mgae}
Chun Wang, Shirui Pan, Guodong Long, Xingquan Zhu, and Jing Jiang.
\newblock Mgae: Marginalized graph autoencoder for graph clustering.
\newblock In \emph{Proceedings of the 2017 ACM on Conference on Information and Knowledge Management}, pages 889--898, 2017.

\bibitem[Yang et~al.(2019)Yang, Huang, Hao, Liu, Belongie, and Hariharan]{1nna-pointflow}
Guandao Yang, Xun Huang, Zekun Hao, Ming-Yu Liu, Serge Belongie, and Bharath Hariharan.
\newblock Pointflow: 3d point cloud generation with continuous normalizing flows.
\newblock In \emph{Proceedings of the IEEE/CVF international conference on computer vision}, pages 4541--4550, 2019.

\end{thebibliography}


\clearpage
\appendix

\end{document}